%% file: main.tex
\documentclass{article} 
\usepackage{iclr2027_conference,times}

\input{math_commands.tex}

\usepackage{hyperref}
\usepackage{algorithm}
\usepackage{amsmath}
\usepackage{colortbl}
\usepackage{pifont}

\definecolor{tabbestbg}{HTML}{FFF0E5}
\definecolor{tabsecondbg}{HTML}{EAF2FF}

\usepackage{pifont}

\usepackage{algpseudocode}
\usepackage[capitalize]{cleveref}
\usepackage{url}
\usepackage{enumitem}
\usepackage{booktabs}
\usepackage{multirow}
\usepackage{graphicx}          
\usepackage[table]{xcolor}     
\definecolor{bestbg}{HTML}{C6DDF2}      
\definecolor{secbg}{HTML}{E9F2FA}       
\newcommand{\best}[1]{\cellcolor{bestbg}\textbf{#1}}
\newcommand{\second}[1]{\cellcolor{secbg}\underline{#1}}
\definecolor{dpos}{HTML}{2E7D32}      
\definecolor{dneg}{HTML}{C62828}      
\definecolor{dzero}{HTML}{8A8A8A}     
\definecolor{bandA}{HTML}{E8EEF8}     
\definecolor{bandB}{HTML}{E3F5F6}     

\newcommand{\codelink}{\url{https://github.com/balibata/PR-OPD}}

\title{PR-OPD: Privileged Representation On-policy Self-Distillation for Agentic Reinforcement Learning}

\author{%
Muyang Li$^{1,*}$, Jie Yang$^{2,*}$, Zhengyu Fang$^{3}$, Junchao Zhu$^{4}$, Zhengkun Xiao$^{1}$, \\
\bfseries Ruining Deng$^{5}$, Zhe Jiang$^{1}$, Shigang Chen$^{1,\dagger}$ \\[4pt]
\mdseries
$^{1}$University of Florida \quad
$^{2}$University of Illinois at Chicago \quad
$^{3}$Case Western Reserve University \\
$^{4}$Vanderbilt University \quad
$^{5}$Weill Cornell Medicine \\[2pt]
\texttt{muyang.li@ufl.edu},\qquad \texttt{jyang265@uic.edu}
}

\iclrfinalcopy 
\begin{document}

\maketitle
\lhead{Preprint.}
{\renewcommand{\thefootnote}{\ensuremath{*}}\footnotetext{Equal contribution. \quad $^{\dagger}$Corresponding author.}}

\begin{abstract}

\input{sections/0_Abstract}
\end{abstract}

\input{sections/1_Introduction.tex}

\input{sections/2_RelatedWorks}

\input{sections/3_Method}

\input{sections/4_Experiments}

\input{sections/6_Conclusion}

\bibliography{main} 

\bibliographystyle{iclr2027_conference}

\appendix
\input{sections/7_Appendix}

\end{document}

%% file: math_commands.tex
\usepackage{amsmath,amsfonts,bm}

\def\eqref#1{equation~\ref{#1}}

\def\1{\bm{1}}

\DeclareMathAlphabet{\mathsfit}{\encodingdefault}{\sfdefault}{m}{sl}
\SetMathAlphabet{\mathsfit}{bold}{\encodingdefault}{\sfdefault}{bx}{n}



%% file: sections/0_Abstract.tex
Language-model agents are usually trained by reinforcement learning from one reward per episode, and privileged self-distillation enriches it by letting the same policy, given a skill, teach its skill-free self through token probabilities.
However, we identify two phenomena that question this channel.
\textbf{Invisible Advantage}: a skill in context lifts WebShop success from 42.2\% to 56.2\%, yet changes the probabilities of fewer than a quarter of the sampled tokens.
\textbf{Much to Align}: a skill changes the hidden states of over 80\% of response tokens, in a way that linear probes can trace back to the specific skill.
To exploit this, we propose Privileged Representation On-policy Self-Distillation~(PR-OPD).
After a GRPO warm start, the policy writes a hindsight skill for each trajectory, re-reads its own responses with that skill as a stop-gradient teacher, and aligns its projected hidden states to the teacher's at every layer alongside the reward objective, with no external skill library, separate teacher, or inference overhead.
On ALFWorld and WebShop with two backbones, PR-OPD achieves the best overall results in every setting, improving over GRPO by up to 4.7 points in ALFWorld success and 14.0 points in WebShop accuracy.
Code is available at \codelink.

%% file: sections/1_Introduction.tex
\section{Introduction}
\label{sec:intro}

Reinforcement learning with verifiable rewards~(RLVR) has emerged as a promising approach for training language-model agents to solve multi-turn tasks~\citep{yao2023react,shao2024deepseekmath,wang2025ragen,feng2025gigpo,jin2025searchr1}.
In many such tasks, the policy receives only a terminal trajectory-level reward: one number per episode, assigned after a sequence of decisions and generated tokens.
Recent methods supplement this coarse feedback with on-policy distillation~(OPD), which provides dense token-level guidance on student-generated trajectories~\citep{hinton2015distilling,agarwal2023gkd,gu2024minillm,lu2025opd,lu2026sdar}.
In these self-distillation methods, the same model serves as a privileged teacher when conditioned on additional information~\citep{vapnik2009lupi,penaloza2026pid}, such as task knowledge or hindsight feedback~\citep{lu2026sdar,yang2026opid,wu2026seed}.
The student learns from the teacher's output probabilities without access to this information, aiming to internalize the guidance so that no additional context is needed at inference time.

These methods rest on an implicit assumption: the teacher's output probabilities provide a sufficiently informative signal for transferring privileged guidance to the student.
However, these output probabilities reflect what the teacher is likely to generate given the additional information, rather than directly revealing how that information changes its internal reasoning process, and a model's internal representations can carry information that its outputs do not express~\citep{gurnee2026verbalizable}.
For example, hindsight feedback may help the teacher identify an overlooked task constraint or an earlier mistake, while producing little change in the probability of a particular sampled token.
In such cases, the token-level probability difference provides a limited signal for transferring the teacher's updated understanding to the student, consistent with recent analyses of the sparsity and dynamics of on-policy distillation signals~\citep{li2026rethinkingopd,yu2026sparsity,shen2026geometry}.
This motivates us to first examine whether privileged information substantially changes the teacher's token probabilities, and then investigate how it affects the teacher's internal processing.

To investigate, we take the natural self-evolving variant of this recipe, in which the policy writes a hindsight skill for each of its own completed trajectories, and compare two forward passes of the same policy over the same sampled responses, one with ordinary context and one with the skill added, as shown in Figure~\ref{fig:diag}.
We also let the policy solve tasks with and without a skill in context, to see whether the skill makes it a better actor.
The skill turns out to make the teacher act better, yet it leaves most of its trace in the model's hidden states rather than in its next-token probabilities.
Our findings highlight two key phenomena:

\begin{figure}[t]
    \centering
    \includegraphics[width=\linewidth]
        {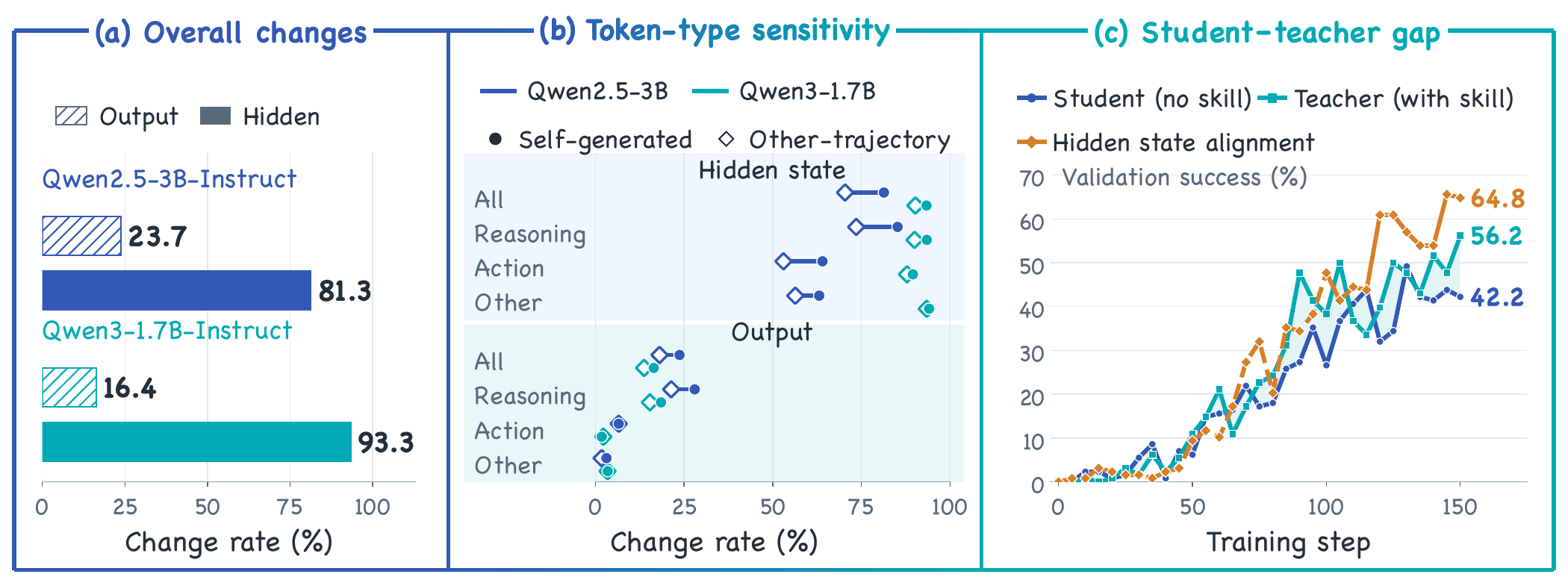}
    \caption{
        \textbf{Two phenomena in privileged self-distillation.}
        \textbf{(a,b)} Fraction of tokens whose hidden states or sampled-token probabilities change when the policy re-reads its own responses with its self-generated skill, on ALFWorld at GRPO@100.
        \textbf{(c)} In-training WebShop validation success on Qwen2.5-3B-Instruct of the student (no skill), the teacher (skill in context), and a student trained by directly aligning to its teacher's hidden states after branching from GRPO at step 100.
    }
    \label{fig:diag}
\end{figure}

\ding{182} \textbf{Invisible Advantage: \emph{the skill makes the teacher act better, yet barely changes its token distribution.}}
As shown in Figure~\ref{fig:diag}(c), a skill in context makes the policy a better actor, lifting WebShop success from 42.2\% to 56.2\%.
Yet Figure~\ref{fig:diag}(a,b) shows that adding the skill leaves the teacher's next-token distribution almost where the student's was: fewer than a quarter of response tokens change their probability, and at action tokens, where decisions are made, almost none do.
The advantage is thus real but nearly invisible in the output channel.
An objective that distills token probabilities can therefore transfer only a thin slice of what the skill gives the teacher.

\ding{183} \textbf{Much to Align: \emph{the skill reshapes the teacher's hidden states, and aligning with them passes the advantage on.}}
As shown in Figure~\ref{fig:diag}(a,b), the same two passes differ in the hidden states of more than 80\% of response tokens, including most action tokens where the output channel is silent.
The change is not noise: linear probes on hidden states can tell a trajectory's own skill from another trajectory's, while probes on output features stay at chance.
Figure~\ref{fig:diag}(c) further shows that this signal is learnable: a student that directly aligns with its teacher's hidden states, with no added parameters, ends up even above the teacher, whereas the student that acts without the skill stays far behind.
What the skill adds lives in how the teacher represents the student's responses, not in its output tokens.

Together, these findings show that the advantage a skill gives the teacher barely reaches its token probabilities, while its hidden states carry it.
This leads to a central question: \emph{\textbf{Can the student acquire the advantage that hindsight gives its privileged self, even though this advantage barely shows in the teacher's tokens?}}

To answer this question, we propose Privileged Representation On-policy Self-Distillation~(\textbf{PR-OPD}).
The key idea is to transfer the teacher's advantage through the channel that actually carries it: instead of imitating the privileged teacher's token probabilities, the student aligns with the teacher's hidden states on the same responses.
Specifically, PR-OPD first trains the policy with GRPO until it can act and analyze its own experience.
It then keeps optimizing the reward and, for each completed trajectory, has the policy write a hindsight skill and re-read its own sampled responses with that skill as a stop-gradient teacher.
The student's hidden states are passed through a lightweight per-layer MLP projector and aligned to the teacher's across all layers.
Student, hindsight analyzer, and teacher are three roles of one set of weights, and the projector is the only added module, so PR-OPD needs no external skill library, no separate teacher model, and nothing extra at inference.

Our main contributions are summarized as follows:
\begin{itemize}[leftmargin=*]
    \item We identify two phenomena in privileged self-distillation, \textbf{Invisible Advantage} and \textbf{Much to Align}: a skill makes the teacher act better yet barely changes its token distribution, and the advantage is carried by its hidden states instead.
    \item We propose \textbf{PR-OPD}, in which the policy learns from its own hindsight by aligning its projected hidden states to those of its privileged self at every layer.
    \item Experiments on ALFWorld and WebShop with two backbones show that PR-OPD achieves the best average success and accuracy in every setting, improving over GRPO by up to $+4.7$ points on ALFWorld and $+11.7$ points on WebShop with Qwen2.5-3B-Instruct.
\end{itemize}

%% file: sections/2_RelatedWorks.tex
\section{Related Work}
\label{sec:related_work}

\subsection{Agentic RL with Privileged Self-Distillation}
Reinforcement learning is the standard way to train language-model agents over multi-turn interaction, with GRPO~\citep{shao2024deepseekmath} and its multi-turn extensions~\citep{wang2025ragen,feng2025gigpo} learning from a sparse terminal reward.
To densify this signal, one line places past experience in the agent's context, such as Reflexion~\citep{shinn2023reflexion}, ExpeL~\citep{zhao2024expel}, and SkillRL~\citep{xia2026skillrl}, and further work trains agents with reinforcement learning to use, internalize, or evolve such skills~\citep{lu2026skill0,shi2026skill1,yao2026skillrise}.
Another line turns it into training-time supervision through privileged on-policy self-distillation, which builds on learning using privileged information~\citep{vapnik2009lupi,lopezpaz2016unifying} and privileged teachers in robot learning~\citep{pinto2018asymmetric,chen2020cheating}, as well as context distillation and on-policy distillation for language models~\citep{snell2022context,agarwal2023gkd,zhao2026opsd}: the same policy, conditioned on a skill it will not see at test time, serves as a teacher whose token log-probabilities are distilled back into the student, as in SDAR~\citep{lu2026sdar}, PCSD~\citep{lv2026pcsd}, OPID~\citep{yang2026opid}, SEED~\citep{wu2026seed}, and related variants~\citep{penaloza2026pid,wang2026agentopsd,pan2026bcsd,tu2026ucob}.
However, all of these read the teacher after the language-model head, so whatever the skill changes inside the teacher reaches the student only through one scalar per token.
To address this, we propose PR-OPD, which aligns the student's hidden states to those of its privileged self.

\subsection{Representation-Level Distillation}
Beyond matching output distributions~\citep{hinton2015distilling,gu2024minillm}, intermediate representations have long served as distillation targets, from hidden-layer hints in FitNets~\citep{romero2015fitnets} to multi-layer transfer in Patient-KD~\citep{sun2019pkd}, TinyBERT~\citep{jiao2020tinybert}, and MiniLM~\citep{wang2020minilm}, all for compressing a fixed, stronger teacher into a smaller student.
More recently, OPRD~\citep{yang2026oprd} aligns hidden states on student-generated responses for mathematical reasoning, and LastOPD~\citep{yang2026lastopdtamingcollapselatent} shows that with a separate, larger teacher this signal collapses late in training because depth-paired layers play different roles in the two models, and therefore restricts alignment to the last layer during a brief crossfade into token-level distillation.
However, these methods assume a teacher that is separate from the student and fixed during training, whereas in self-distillation for agents the teacher is the policy itself, so its layers correspond one-to-one with the student's, its hindsight is written by that policy, and both change at every update.
None of them asks the question raised by our diagnostics: whether an advantage that barely shows in the teacher's tokens can still be passed on through its hidden states.
This motivates PR-OPD, which treats the policy under its own hindsight as a stop-gradient teacher and aligns the student's hidden states to it across all layers inside the RL loop, so that the student learns from the teacher's hidden states rather than imitating its token probabilities.

%% file: sections/3_Method.tex
\section{Method}
\label{sec:method}

\begin{figure}[t]
    \centering
    \includegraphics[width=0.99\linewidth]{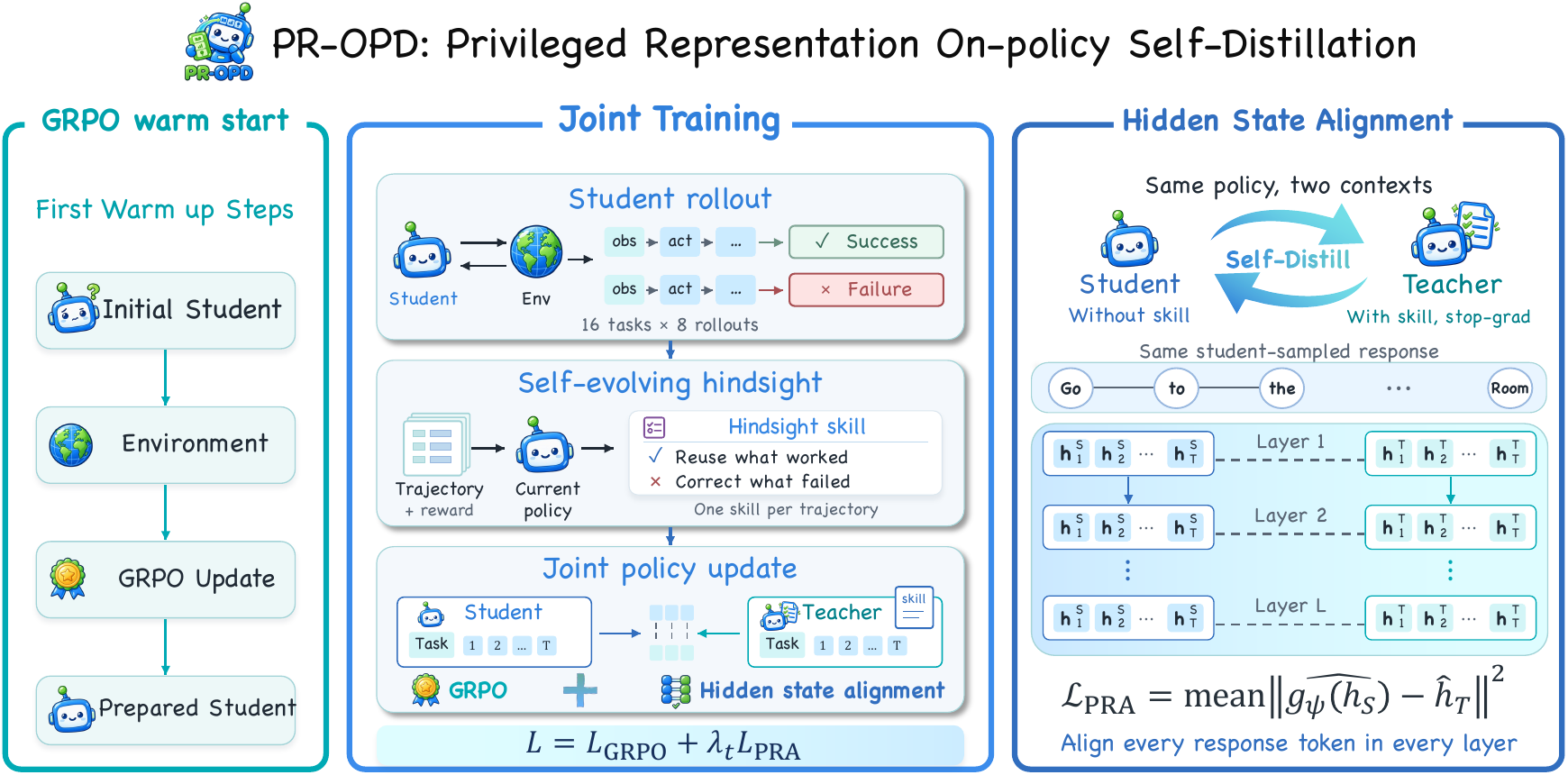}
    \caption{\textbf{Overview of PR-OPD.}
    \textbf{Left:} a GRPO warm start first makes the policy able to act and to analyze its own experience.
    \textbf{Center:} in each later iteration, the policy collects trajectories, writes one hindsight skill per completed trajectory, and updates with the GRPO loss plus a hidden-state alignment loss.
    \textbf{Right:} the same policy re-reads the student-sampled response twice, without and with the skill. The skill-conditioned pass is the stop-gradient teacher, and every response token in every layer of the student is passed through a lightweight MLP projector and pulled toward it.
    Skills and teacher states are recomputed as the policy evolves, and neither is needed at inference.}
    \label{fig:framework}
\end{figure}

\subsection{Problem Definition}
\label{sec:problem}

Consider a policy $\pi_\theta$ that solves a task $q$ over multiple turns: for each of $G$ sampled trajectories, $\tau_i=\{(x_{i,u},y_{i,u})\}_{u=1}^{U_i}$ records the context $x_{i,u}$ and the sampled response $y_{i,u}$ at every turn $u$, and the whole trajectory receives one terminal reward $R_i$.
This context holds the task, observations, and interaction history, and is all the policy ever sees at inference.
During training, however, the same policy can take on two further roles once a trajectory is complete: as the \emph{analyzer} it reads $(\tau_i,R_i)$ and writes a hindsight skill $s_i$, and as the \emph{teacher} it re-reads the student's own responses $y_{i,u}$ with $s_i$ prepended to the context.
The catch, as Figure~\ref{fig:diag} shows, is that the skill makes this teacher act better yet barely changes its token probabilities~(Invisible Advantage), while its hidden states change broadly and carry the advantage~(Much to Align).
Our goal is therefore to pass this advantage to the student through the teacher's hidden states rather than its token probabilities, without giving up reward learning or adding anything at inference.

\subsection{Overview}
\label{sec:method-overview}

PR-OPD runs in two stages~(Figure~\ref{fig:framework}).
Stage~I trains the policy with GRPO alone for $N$ updates.
This warm start is needed because the analyzer is the policy itself, and it has to learn the task before it can explain its own mistakes.
Stage~II keeps the reward objective and adds representation supervision.
Each iteration collects trajectories without skills, has the current policy write one skill per completed trajectory, and updates the policy with the GRPO loss plus an alignment loss toward the teacher's hidden states.
Student, analyzer, and teacher share one set of weights throughout, and the only added module is a lightweight per-layer MLP projector on the student side, trained jointly with the policy and discarded at inference.

\subsection{Stage I: GRPO Warm Start}
\label{sec:warmstart}

For each task $q$, the rollout policy $\pi_{\theta_{\mathrm{old}}}$ samples $G$ trajectories with terminal rewards $\mathbf{R}=\{R_i\}_{i=1}^{G}$, and GRPO~\citep{shao2024deepseekmath} gives every response token of trajectory $i$ the group-relative advantage $A_i=(R_i-\operatorname{mean}(\mathbf{R}))/\operatorname{std}(\mathbf{R})$.
With the ratio $r_{i,k}(\theta)=\pi_\theta(y_{i,k}\mid\cdot)/\pi_{\theta_{\mathrm{old}}}(y_{i,k}\mid\cdot)$ for the $k$-th of the $K_i$ response tokens in $\tau_i$, the policy minimizes
\begin{equation}
\mathcal{L}_{\mathrm{GRPO}}(\theta)
=-\frac{1}{G}\sum_{i=1}^{G}\frac{1}{K_i}\sum_{k=1}^{K_i}
\min\Bigl(r_{i,k}(\theta)A_i,\ \operatorname{clip}\bigl(r_{i,k}(\theta),1-\epsilon,1+\epsilon\bigr)A_i\Bigr).
\label{eq:grpo}
\end{equation}
Stage~I optimizes this objective alone for $N$ updates, without skills or teacher passes, so that the hindsight the policy later writes about its own trajectories is worth learning from.

\subsection{Self-Generated Hindsight Skill}
\label{sec:hindsight-teacher}

\paragraph{Generate one skill per trajectory.}
At Stage-II iteration $j$, the rollout policy $\pi_{\theta_j}$ collects fresh trajectories without privileged information.
After an episode finishes, we serialize its interaction record and outcome into a compact text $z_i$ and prompt the same policy to write a hindsight skill:
\begin{equation}
z_i=\operatorname{Serialize}(\tau_i,R_i),
\qquad
s_i\sim\pi_{\theta_j}(\,\cdot\mid z_i,I_{\mathrm{hind}}),
\label{eq:hindsight-skill}
\end{equation}
where the instruction $I_{\mathrm{hind}}$ asks for a short list of actionable rules.
For a successful trajectory, these rules describe a reusable workflow and the observations that decided it, and for a failed one, they name the critical mistakes and how to correct them.
Skill generation is non-differentiable, and the analyzer receives no separate SFT stage.
Skills are rewritten from every new rollout batch rather than retrieved from a library.

\paragraph{Revisit the same responses with hindsight.}
For every turn of trajectory $i$, we prepend its skill to the turn's context $x$ with a fixed template:
\begin{equation}
\widetilde{x}=H(x,s_i).
\label{eq:teacher-context}
\end{equation}
The student processes $(x,y)$ and the teacher processes $(\widetilde{x},y)$, so both passes see the same response tokens and the teacher generates no replacement actions.
The skill is privileged because it comes from the completed episode, including observations and outcomes unavailable at decision time.

\subsection{Privileged Representation Alignment}
\label{sec:propd}

\paragraph{Shared weights and matched positions.}
Following the representation-level formulation of OPRD~\citep{yang2026oprd}, we extract hidden states along the student-sampled responses.
To keep notation light, we consider one turn with context $x$, response $y$, and teacher context $\widetilde{x}$, and drop the trajectory and turn indices.
For layer $\ell$ and response token $y_k$, define
\begin{equation}
h^{(\ell)}_{S,k}=f^{(\ell)}_{\theta}(x,y_{<k}),
\qquad
h^{(\ell)}_{T,k}=f^{(\ell)}_{\theta}(\widetilde{x},y_{<k}),
\label{eq:paired-states}
\end{equation}
where $f^{(\ell)}_{\theta}$ returns the layer-$\ell$ hidden state just before token $y_k$ is predicted.
The two passes run the same network with the same current weights $\theta$ and differ only in their prompts.
Layer $\ell$ of the student is therefore paired with layer $\ell$ of the teacher, with no layer mapping to choose as in distillation from a different model~\citep{yang2026lastopdtamingcollapselatent}.
Because the teacher's prompt is longer by the skill, the same response token sits at a different absolute position in each pass, so we pair states by response token rather than by position.
Within Stage-II iteration $j$, $\theta$ starts from $\theta_j$, the weights of the rollout policy that collected the batch, and changes as the batch is optimized.

\paragraph{Align every layer and every response token.}
Let $\mathcal{M}$ collect the valid response tokens of all turns in a training microbatch.
Each student state first passes through a per-layer MLP projector $g^{(\ell)}_{\psi}:\mathbb{R}^{d}\to\mathbb{R}^{d}$ with one 512-dimensional hidden layer, giving $p^{(\ell)}_{S,k}=g^{(\ell)}_{\psi}(h^{(\ell)}_{S,k})$.
Since the student's raw states also feed its own language-model head, the projector lets it follow the teacher without forcing those states to copy ones computed under a longer, skill-bearing context.
With $\widehat{v}=v/\max(\|v\|_2,\varepsilon)$, we minimize the privileged representation alignment~(PRA) loss
\begin{equation}
\mathcal{L}_{\mathrm{PRA}}(\theta,\psi)
=
\frac{1}{L|\mathcal{M}|}
\sum_{\ell=1}^{L}
\sum_{k\in\mathcal{M}}
\left\|
\widehat{p}^{(\ell)}_{S,k}
-
\operatorname{sg}\!\left[
\widehat{h}^{(\ell)}_{T,k}
\right]
\right\|_2^2,
\label{eq:propd}
\end{equation}
where $L$ counts all Transformer layers excluding the embedding output and $\operatorname{sg}$ stops gradients through the teacher.
$\mathcal{M}$ includes reasoning and action tokens in the response but not prompt or padding.
Whenever both norms exceed $\varepsilon$, each normalized distance equals $2-2\cos(p_S,h_T)$~(Appendix~\ref{app:cosine}), so alignment concerns direction rather than magnitude.
Because the teacher shares the current weights, its states are recomputed at every update and the target moves with the policy.

\subsection{Overall Objective}
\label{sec:propd-objective}

The two stages share one objective, in which the alignment loss is added to the RL loss only after the $N$ warm-start updates:
\begin{equation}
\mathcal{L}(\theta,\psi;t)
=
\underbrace{
\mathcal{L}_{\mathrm{GRPO}}(\theta)
+\beta\mathcal{L}_{\mathrm{KL}}(\theta)
-\eta\mathcal{H}(\pi_\theta)
}_{\mathcal{L}_{\mathrm{RL}}(\theta)}
+\lambda_t\mathcal{L}_{\mathrm{PRA}}(\theta,\psi),
\qquad
\lambda_t=
\begin{cases}
0, & t\leq N,\\
\lambda, & t>N,
\end{cases}
\label{eq:total}
\end{equation}
where $\mathcal{L}_{\mathrm{KL}}$ regularizes the student toward a frozen reference policy and $\mathcal{H}$ is the policy entropy.
No output-space distillation term is used.
Algorithm~\ref{alg:propd} in Appendix~\ref{app:algorithm} summarizes one Stage-II iteration.
After each update, the new policy acts, writes skills, and serves as teacher in the next iteration, so the experience and its hindsight supervision evolve together.

%% file: sections/4_Experiments.tex
\section{Experiments}
\label{sec:exp}

\input{tables/main_performance}

\subsection{Experimental Settings}
\label{sec:setup}

\textbf{Benchmarks.}
We evaluate on ALFWorld~\citep{shridhar2021alfworld}, a text-based household environment, and WebShop~\citep{yao2022webshop}, a web-shopping environment.
We report success rates on ALFWorld and task accuracy and graded score on WebShop, with split details in Appendix~\ref{app:setup}.

\textbf{Protocol.}
We use Qwen2.5-3B-Instruct~\citep{yang2024qwen25} and Qwen3-1.7B-Instruct~\citep{yang2025qwen3} as backbones.
Every method trains for 150 updates, and PR-OPD spends the first $N$ of them on GRPO, with $N=120$ on WebShop and $N=100$ on ALFWorld.
All methods are evaluated at step 150 on 128 episodes with hyperparameters and hardware in Appendix~\ref{app:setup}.

\textbf{Baselines.}
We compare against six baselines that each test one alternative explanation: Vanilla and Skill-Prompt whether a skill helps the untrained policy, GRPO~\citep{shao2024deepseekmath} whether reinforcement learning alone suffices, SDAR~\citep{lu2026sdar} and OPSD~\citep{zhao2026opsd} whether distilling token probabilities is enough, and Skill-SD~\citep{wang2026skillsd} whether a stronger token-level objective closes the gap.
All skill-based baselines use the hand-written skills of SDAR~\citep{lu2026sdar}, one per task type.

\subsection{Main Results}
\label{sec:main_results}

We compare PR-OPD against the six baselines on ALFWorld and WebShop under both backbones.
Results are in Table~\ref{tab:main}.
Based on these results, we summarize our observations~(\textbf{Obs.}) as follows:


\textbf{Obs.~\ding{182}: PR-OPD achieves the best aggregate result under every backbone.}
It attains the highest ALFWorld overall success and the highest WebShop accuracy and score under both backbones.
On Qwen2.5-3B-Instruct, it improves over GRPO by 4.7 points on ALFWorld and by 11.7 points in WebShop accuracy.
It also ranks first or second on at least four of the six ALFWorld task types under each backbone.
These results answer the central question of Section~\ref{sec:intro}: the student can acquire the advantage that hindsight gives its privileged self, without seeing the skill at inference.

\textbf{Obs.~\ding{183}: Aligning hidden states with self-written skills beats distilling tokens with curated ones.}
SDAR, OPSD, and Skill-SD all distill the teacher's token probabilities under skills from the same curated library, matched by task type, whereas PR-OPD aligns hidden states under a skill it writes for each trajectory.
Against the strongest of them, Skill-SD, PR-OPD gains 7.8 points in WebShop accuracy and 8.1 in score with the 3B backbone, and it also leads on ALFWorld under both backbones.
The other token-level methods gain little: SDAR improves over GRPO by at most 2.4 points on ALFWorld, and OPSD falls to roughly the level of the untrained policy.
This matches the \textbf{Much to Align} of Phenomenon~\ding{183}: the skill's advantage is carried by the teacher's hidden states, so aligning them transfers more than distilling its tokens.

\subsection{Ablation and Sensitivity Study}
\label{sec:ablation}
\label{sec:core_ablation}
\label{sec:sensitivity}
\label{sec:warm_start_analysis}

\begin{figure}[t]
    \centering
    \includegraphics[width=\linewidth]{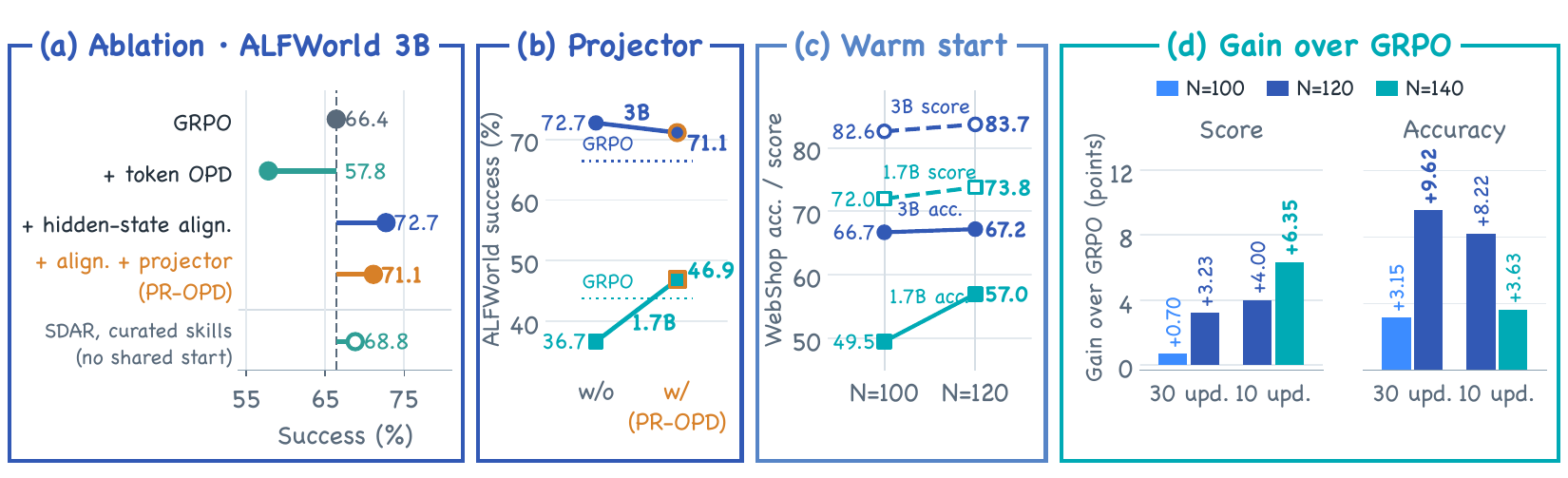}
    \caption{\textbf{Ablation and sensitivity study.}
    \textbf{(a)} ALFWorld success of each variant with the 3B backbone, relative to GRPO (dashed line).
    \textbf{(b)} ALFWorld success without and with the projector for both backbones, with GRPO as dotted lines.
    \textbf{(c)} WebShop accuracy (filled) and score (open) of PR-OPD started at $N=100$ or $N=120$.
    \textbf{(d)} Gain of direct alignment over GRPO continued from the same checkpoint on WebShop with 1.7B, averaged over six seeds.
    Numbers are in Appendix~\ref{app:ablation_tables}.}
    \label{fig:ablation}
\end{figure}

We examine each design choice of PR-OPD by changing one at a time, named as in Figure~\ref{fig:ablation}.
Ablations use ALFWorld, with the projector compared on both backbones, and the warm-start study uses WebShop with the 1.7B backbone.
(\textbf{1})~\textbf{+ token OPD} distills the teacher's token probabilities under the same self-generated skills.
(\textbf{2})~\textbf{+ hidden-state align.} aligns hidden states directly, without the projector.
(\textbf{3})~\textbf{+ align. + projector} is the full PR-OPD.
(\textbf{4})~\textbf{SDAR} distills token probabilities under curated skills and trains from scratch.
(\textbf{5})~\textbf{$N$} starts representation supervision after 100, 120, or 140 GRPO updates, and for direct alignment we pair each start with GRPO continued from the same checkpoint over six evaluation seeds.
Variants (1) to (3) branch from the GRPO checkpoint at step 100, and all runs end at step 150.
From Figure~\ref{fig:ablation}, we observe:

\textbf{Obs.~\ding{184}: Reading the teacher's hidden states, not its tokens, drives the gain.}
From the same warm start and the same self-generated skills, token-level OPD falls to 57.8\%, 8.6 points below GRPO, whereas hidden-state alignment rises to 72.7\%, 6.3 points above (Figure~\ref{fig:ablation}a).
The two runs differ only in how the teacher is read, yet they end 14.9 points apart.
SDAR, which reads tokens under curated skills, gains only 2.4 points.
This is the \textbf{Invisible Advantage} of Phenomenon~\ding{182} under controlled conditions: the teacher holds the advantage, but its tokens cannot pass it on.

\textbf{Obs.~\ding{185}: The projector is nearly free on the larger backbone and essential on the smaller one.}
With the 3B backbone, the projector moves success from 72.7\% to 71.1\%.
With the 1.7B backbone, where direct alignment falls to 36.7\%, below GRPO, the projector lifts ALFWorld success to 46.9\%, the best result in Table~\ref{tab:main} (Figure~\ref{fig:ablation}b).
A smaller policy thus benefits from a learned buffer between its own states and the teacher's, which is why PR-OPD includes the projector by default~(Appendix~\ref{app:align_dynamics}).

\textbf{Obs.~\ding{186}: The starting point matters more than the number of supervised updates.}
Starting at $N=120$ instead of $N=100$ raises the WebShop accuracy of PR-OPD from 49.5\% to 57.0\% with 20 fewer supervised updates (Figure~\ref{fig:ablation}c).
Paired against GRPO continued from the same checkpoint over six evaluation seeds, direct alignment started at step 120 gains 9.62 points after 30 updates and 8.22 after 10, against 3.15 from step 100 and 3.63 from step 140 (Figure~\ref{fig:ablation}d).
The policy must first learn the task before its own hindsight is worth reading, hence the GRPO warm start.

\subsection{General Analysis}
\label{sec:general_analysis}
\label{sec:ablation_analysis}
\label{sec:layer_sensitivity}

\begin{figure}[t]
    \centering
    \includegraphics[width=\linewidth]{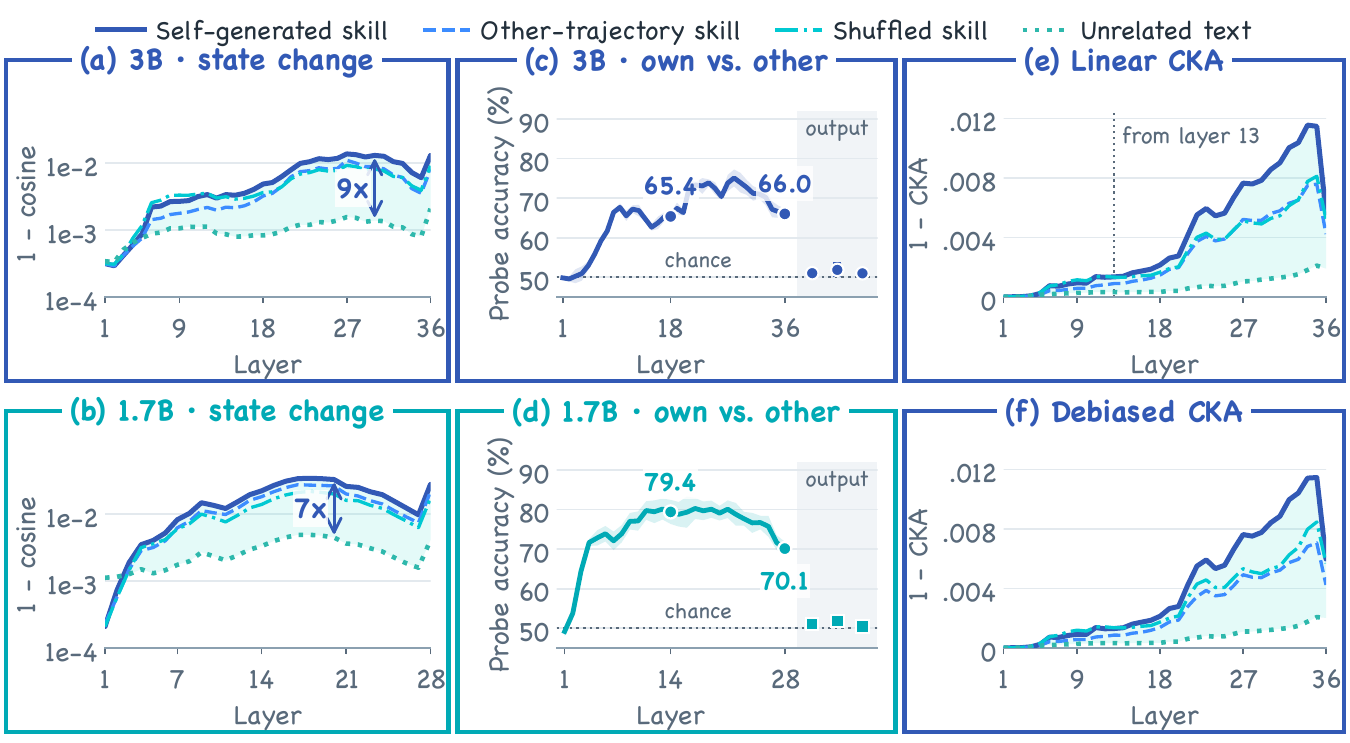}
    \caption{\textbf{How the skill acts on the teacher's hidden states across depth.}
    Columns show three analyses at fixed checkpoints on ALFWorld, with 3B on top and 1.7B below in (a--d).
    \textbf{(a,b)} Mean cosine distance from the skill-free context at each layer at GRPO@100, on a logarithmic axis.
    \textbf{(c,d)} Probe accuracy for the trajectory's own versus another trajectory's skill at each layer, with output features at the right.
    \textbf{(e,f)} Same-layer $1-\mathrm{CKA}$ of the initial 3B policy under linear and debiased CKA.
    In (a,b,e,f), shaded bands mark the gap between the own skill and unrelated text.}
    \label{fig:depth}
\end{figure}

\begin{figure}[t]
    \centering
    \includegraphics[width=\linewidth]{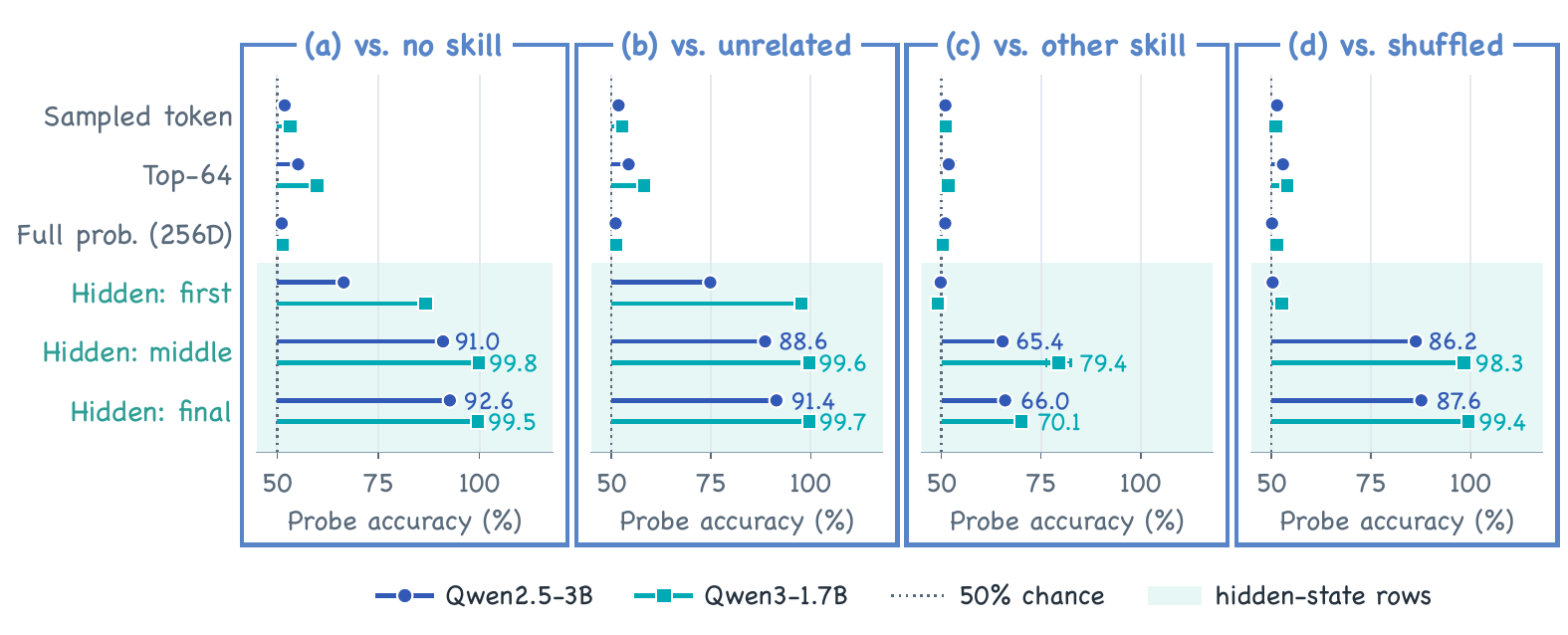}
    \caption{\textbf{Probing the hindsight context at selected layers.}
    Linear-probe accuracy at GRPO@100 on ALFWorld for distinguishing a trajectory's own skill from (a) no skill, (b) unrelated text, (c) another trajectory's skill, and (d) a token-shuffled skill.
    The first three rows are output features read after the language-model head, and the shaded rows are hidden states.}
    \label{fig:hindsight_decodability}
\end{figure}

We next examine where the skill acts inside the teacher and what it carries.
At fixed checkpoints on ALFWorld, we fix weights and responses and vary the context among four conditions: the trajectory's own skill, another trajectory's skill, a token-shuffled skill, and length-matched unrelated text.
We measure how much each context changes hidden states at every layer, probe~\citep{alain2016probes,belinkov2022probing} whether hidden or output features tell the contexts apart, and compare hidden-state structure with linear CKA~\citep{kornblith2019cka}.
From Figures~\ref{fig:depth} and~\ref{fig:hindsight_decodability}, we observe:

\textbf{Obs.~\ding{187}: Hindsight reshapes hidden states across the whole network, far beyond what unrelated text does.}
In both backbones, the hidden-state change grows from early to middle layers and stays high through the later layers, rather than concentrating next to the output head (Figure~\ref{fig:depth}a,b).
All three skill contexts move the states well above length-matched unrelated text, by up to nine times for 3B and seven times for 1.7B, and the trajectory's own skill moves them the most.
The change is therefore specific to the skill rather than to a longer prompt.

\textbf{Obs.~\ding{188}: Hidden states carry which skill was given, while output features do not.}
Linear probes on middle-layer hidden states separate a trajectory's own skill from every alternative in Figure~\ref{fig:hindsight_decodability}, exceeding 88\% against no skill or unrelated text, and reaching 65.4\% and 79.4\% against another trajectory's skill and 86.2\% and 98.3\% against a token-shuffled copy for 3B and 1.7B~(Appendix~\ref{app:probes}).
The three output features stay within 50.2--54.0\% on the last two distinctions, close to chance.
The information the skill adds is thus readable from the teacher's hidden states, not its outputs.

\textbf{Obs.~\ding{189}: The skill information peaks in the middle layers and persists to the output.}
Probing every layer for own versus other-trajectory skills, accuracy starts near chance at the first layer, rises through the early layers, and stays well above the output features through the rest of the network~(Figure~\ref{fig:depth}c,d and Appendix~\ref{app:layerwise_decodability}).
For 1.7B, accuracy peaks at 79.4\% in the middle layers and drops to 70.1\% at the final layer, and for 3B the final layer also sits below the intermediate peak, while all three output features stay at 50--52\%.
Aligning only the last layer as LastOPD does~\citep{yang2026lastopdtamingcollapselatent} would therefore miss where this information is clearest, so PR-OPD aligns every layer.

\textbf{Obs.~\ding{190}: The skill makes targeted adjustments to the structure of the hidden states.}
Unlike cosine distance, linear CKA compares how response positions relate to each other within a layer, so it captures structure rather than the direction of each vector.
On the initial 3B policy, before any warm-start training, the trajectory's own skill leaves the overall structure largely intact, with same-layer CKA averaging 0.996, yet from layer 13 onward it changes the structure more than all three controls~(Figure~\ref{fig:depth}e and Appendix~\ref{app:cka_hindsight}).
Averaged across layers, the own skill changes the structure 1.5 times as much as another trajectory's skill and six times as much as unrelated text, and the debiased estimate gives the same ordering (Figure~\ref{fig:depth}f).
The skill therefore reshapes the representation in a targeted, depth-dependent way rather than overwriting it, even before any training.
These analyses place the skill's advantage in the teacher's hidden states at every layer, exactly what PR-OPD aligns.

%% file: tables/main_performance.tex
\begin{table}[t]
\centering
\caption{Step-150 results on ALFWorld and WebShop.
ALFWorld reports success rates (\%), and WebShop reports accuracy (Acc, \%) and score.
Within each backbone, \colorbox{bestbg}{\textbf{bold}} is the best and \colorbox{secbg}{\underline{underline}} the second best.}
\label{tab:main}
\small
\setlength{\tabcolsep}{6pt}
\begin{tabular}{l|ccccccc|cc}
\toprule
& \multicolumn{7}{c|}{\textbf{ALFWorld}} & \multicolumn{2}{c}{\textbf{WebShop}} \\
\textbf{Method} & Pick & Look & Clean & Heat & Cool & Pick2 & All & Acc & Score \\
\midrule
\rowcolor{bandA}\multicolumn{10}{c}{\textbf{Qwen2.5-3B-Instruct}} \\
Vanilla & 44.4 & 28.6 & 25.0 & 7.7 & 14.3 & 4.3 & 23.4 & 0.8 & 5.4 \\
Skill-Prompt & 44.4 & \second{63.6} & 29.2 & 5.3 & 4.8 & 5.9 & 25.8 & 0.8 & 3.1 \\
GRPO & 88.9 & \best{72.7} & 50.0 & 52.6 & \best{61.9} & \best{58.8} & 66.4 & 55.5 & 70.4 \\
SDAR & 88.9 & \second{63.6} & \second{66.7} & \best{63.2} & \best{61.9} & 47.1 & 68.8 & 56.3 & 74.7 \\
OPSD & 55.6 & 35.7 & 11.1 & 0.0 & 14.3 & 20.0 & 23.4 & 1.6 & 6.0 \\
Skill-SD & \second{91.9} & \best{72.7} & \best{70.8} & 55.6 & \second{57.1} & 47.1 & \second{69.5} & \second{59.4} & \second{75.6} \\
\textbf{PR-OPD}~(ours) & \best{97.2} & 54.5 & \best{70.8} & \second{57.9} & \best{61.9} & \second{52.9} & \best{71.1} & \best{67.2} & \best{83.7} \\
\midrule
\rowcolor{bandB}\multicolumn{10}{c}{\textbf{Qwen3-1.7B-Instruct}} \\
Vanilla & 18.9 & 42.9 & 0.0 & 0.0 & 4.8 & 0.0 & 8.6 & 3.1 & 44.7 \\
Skill-Prompt & 5.6 & \best{63.6} & 41.7 & 5.3 & 0.0 & 5.9 & 16.4 & 2.3 & 25.7 \\
GRPO & \second{58.3} & 45.5 & \second{50.0} & 36.8 & \second{23.8} & \best{35.3} & 43.8 & 43.0 & 66.4 \\
SDAR & \second{58.3} & \best{63.6} & 45.8 & \second{47.4} & \second{23.8} & \best{35.3} & \second{46.1} & 45.3 & 66.6 \\
OPSD & 25.0 & \second{54.5} & 8.3 & 0.0 & 4.8 & 5.9 & 14.8 & 6.3 & 44.2 \\
Skill-SD & 55.6 & 45.5 & \best{58.3} & \best{52.6} & \best{28.6} & \second{17.6} & 45.3 & \second{54.7} & \second{73.7} \\
\textbf{PR-OPD}~(ours) & \best{69.4} & \best{63.6} & 37.5 & \best{52.6} & \best{28.6} & \second{17.6} & \best{46.9} & \best{57.0} & \best{73.8} \\
\bottomrule
\end{tabular}
\end{table}

%% file: sections/6_Conclusion.tex
\section{Conclusion}
\label{sec:conclusion}

This paper studies two phenomena of privileged self-distillation for language-model agents: a skill makes the teacher act better yet barely changes its token probabilities, and the advantage it brings is carried instead by the teacher's hidden states.
We therefore introduce PR-OPD, which lets the policy write a hindsight skill for each trajectory and, after a GRPO warm start, align its projected hidden states to those of its privileged self at every layer.
On ALFWorld and WebShop with two backbones, PR-OPD achieves the best overall results in every setting, and our analyses locate the skill's effect in the teacher's hidden states across all layers rather than in its outputs.
What a policy learns from its own hindsight is thus better read from what its privileged self represents than from what it says.

%% file: sections/7_Appendix.tex
\section{Normalized Alignment Distance}
\label{app:cosine}
For any $a,b\in\mathbb{R}^{d}$ with $\|a\|_2\geq\varepsilon$ and $\|b\|_2\geq\varepsilon$, the normalization in Eq.~\ref{eq:propd} yields the unit vectors $\widehat{a}=a/\|a\|_2$ and $\widehat{b}=b/\|b\|_2$, so
\begin{equation}
\|\widehat{a}-\widehat{b}\|_2^2
=\|\widehat{a}\|_2^2+\|\widehat{b}\|_2^2-2\,\widehat{a}^{\top}\widehat{b}
=2-2\,\frac{a^{\top}b}{\|a\|_2\,\|b\|_2}
=2-2\cos(a,b).
\end{equation}
Setting $a=p^{(\ell)}_{S,k}$ and $b=h^{(\ell)}_{T,k}$ shows that each term of Eq.~\ref{eq:propd} depends only on the angle between the projected student state and the teacher state, and lies in $[0,4]$.
If a norm falls below $\varepsilon$, that vector is divided by $\varepsilon$ instead and its norm stays below one, which is why the identity is stated only for norms of at least $\varepsilon$.

\section{Training Procedure}
\label{app:algorithm}
Algorithm~\ref{alg:propd} lists one Stage-II iteration of PR-OPD.

\begin{algorithm}[!htbp]
\caption{One Stage-II iteration of PR-OPD}
\label{alg:propd}
\small
\begin{algorithmic}[1]
\Require Rollout policy $\pi_{\theta_j}$, task batch $\mathcal{Q}$,
    reference policy $\pi_{\mathrm{ref}}$
\Ensure Updated policy $\pi_{\theta_{j+1}}$
\State Collect grouped trajectories $\mathcal{D}_j$ and rewards
    using $\pi_{\theta_j}$ without skills
\State Cache rollout log-probabilities and compute GRPO advantages
\ForAll{completed trajectories $\tau_i\in\mathcal{D}_j$}
    \State Generate $s_i$ from $(\tau_i,R_i)$ with $\pi_{\theta_j}$
        \Comment{no gradient}
    \State Set $\widetilde{x}=H(x,s_i)$ for every turn of $\tau_i$;
        keep $y$ unchanged
\EndFor
\State $\theta\gets\theta_j$
\ForAll{optimization minibatches $\mathcal{B}\subset\mathcal{D}_j$}
    \State Run the student on $(x,y)$ with current $\theta$;
        retain gradients
    \State Run the teacher on $(\widetilde{x},y)$ with the same $\theta$;
        disable gradients
    \State Compute $\mathcal{L}_{\mathrm{RL}}
        +\lambda\mathcal{L}_{\mathrm{PRA}}$
        using Eqs.~\ref{eq:propd} and~\ref{eq:total}
    \State Accumulate microbatch gradients for $\theta$ and $\psi$ and take an optimizer step
\EndFor
\State $\theta_{j+1}\gets\theta$; synchronize the rollout policy
\end{algorithmic}
\end{algorithm}

\section{Experimental Details}
\label{app:setup}

\textbf{Benchmarks.}
ALFWorld is a text-based embodied environment in which agents follow natural-language instructions to complete multi-step household tasks.
We use its \texttt{eval\_in\_distribution} (seen) split with 140 task instances and report per-task-type and overall success rates.
WebShop requires agents to search, inspect, and purchase products that satisfy user-specified constraints.
We use the standard test split of the first 500 instructions and report task accuracy, which requires full completion, and graded task score, which also credits partial satisfaction of the requirements.

\textbf{Implementation.}
Both backbones have hidden size $d=2048$.
PR-OPD starts from a GRPO checkpoint after $N$ warm-start updates, with $N=100$ on ALFWorld and $N=120$ on WebShop, and continues to global step 150.
The per-layer projector has one 512-dimensional hidden layer and is trained jointly with the policy at a learning rate of $10^{-4}$.
We keep the GRPO baseline's group-relative advantages, clipping, and invalid-action penalty, with KL coefficient $\beta=0.01$, entropy coefficient $\eta=0.001$, and alignment weight $\lambda=0.01$.
Our implementation builds on SDAR~\citep{lu2026sdar} and uses \texttt{verl}~\citep{sheng2025hybridflow} for distributed training and \texttt{vLLM}~\citep{kwon2023vllm} for rollout generation.
All experiments run on eight NVIDIA B200 GPUs with 64 CPU cores and 450~GB of host memory.

\textbf{Evaluation.}
We evaluate the checkpoint at global step 150 with sampling temperature $0.4$ on 128 episodes drawn from the evaluation split.
Each configuration uses a single training run with training seed $0$.
Training curves, as in Figure~\ref{fig:diag}(c), report in-training validation every five steps, whereas all tables use the step-150 evaluation.

\section{Ablation and Sensitivity Results}
\label{app:ablation_tables}

Tables~\ref{tab:ablation} and~\ref{tab:sensitivity} list the numbers shown in Figure~\ref{fig:ablation}.

\begin{table}[t]
\centering
\caption{Ablation on ALFWorld with the 3B backbone, and the projector with the 1.7B backbone at the bottom.
All runs end at step 150.
$\Delta$ is the difference from GRPO in points.}
\label{tab:ablation}
\small
\begin{tabular}{@{}llcr@{}}
\toprule
\textbf{Variant} & \textbf{Teacher channel} & \textbf{Success (\%)} & $\Delta$ \\
\midrule
GRPO & none & 66.4 & 0.0 \\
\quad + token-level OPD & token probabilities & 57.8 & $-8.6$ \\
\quad + hidden-state alignment, w/o projector & hidden states & \textbf{72.7} & $+6.3$ \\
\quad + hidden-state alignment, w/ projector (\textbf{PR-OPD}) & hidden states & 71.1 & $+4.7$ \\
\midrule
SDAR, with curated skills & token probabilities & 68.8 & $+2.4$ \\
\midrule
\multicolumn{4}{@{}l}{\textit{Projector with the 1.7B backbone}} \\
GRPO & none & 43.8 & 0.0 \\
\quad + hidden-state alignment, w/o projector & hidden states & 36.7 & $-7.1$ \\
\quad + hidden-state alignment, w/ projector (\textbf{PR-OPD}) & hidden states & \textbf{46.9} & $+3.1$ \\
\bottomrule
\end{tabular}
\end{table}

\begin{table}[t]
\centering
\caption{Sensitivity to the warm-start point on WebShop with the 1.7B backbone.
Top: PR-OPD at step 150.
Bottom: gain of direct alignment over GRPO continued from the same checkpoint, averaged over six evaluation seeds.}
\label{tab:sensitivity}
\small
\begin{tabular}{@{}cccc@{}}
\toprule
\textbf{Start $N$} & \textbf{Stage-II updates} & \textbf{Acc (\%)} & \textbf{Score} \\
\midrule
100 & 50 & 49.5 & 72.0 \\
120 & 30 & \textbf{57.0} & \textbf{73.8} \\
\midrule
\textbf{Start $N$} & \textbf{Stage-II updates} & \textbf{$\Delta$Acc} & \textbf{$\Delta$Score} \\
\midrule
100 & 30 & $+3.15$ & $+0.70$ \\
120 & 30 & $\mathbf{+9.62}$ & $+3.23$ \\
\midrule
120 & 10 & $+8.22$ & $+4.00$ \\
140 & 10 & $+3.63$ & $\mathbf{+6.35}$ \\
\bottomrule
\end{tabular}
\end{table}

\section{Linear Probing of Hindsight}
\label{app:probes}
\label{sec:hindsight_decodability}

\textbf{Setup.}
At fixed GRPO@100 checkpoints on ALFWorld, we train logistic regression probes to distinguish a trajectory's own self-generated skill from four alternatives: no skill, unrelated filler text, another trajectory's skill, and a token-shuffled version of the same skill.
Evaluation uses five folds grouped by trajectory, so all examples from one trajectory stay in the same fold, and chance accuracy is 50\%.
We compare three output-side features, namely the sampled token's log-probability, the log-probabilities of the skill-free distribution's top-64 candidates, and a 256-dimensional random projection of the full output probability vector, against hidden states from the first, middle, and final layers.
The middle layers are 18 for Qwen2.5-3B-Instruct and 14 for Qwen3-1.7B-Instruct.

\textbf{Results.}
Against another trajectory's skill (Figure~\ref{fig:hindsight_decodability}c), middle-layer hidden states reach 65.4\% and 79.4\% accuracy for 3B and 1.7B, compared with 50.4--51.9\% across the output features, while first-layer hidden states stay near chance at 49.8\% and 49.2\%.
Against a token-shuffled copy (Figure~\ref{fig:hindsight_decodability}d), which keeps the skill's tokens but changes their order, middle-layer probes reach 86.2\% and 98.3\% and final-layer probes 87.6\% and 99.4\%, compared with 50.2--54.0\% for the output features.
Against no skill and unrelated text (Figure~\ref{fig:hindsight_decodability}a,b), middle-layer accuracies range from 88.6\% to 99.8\%, compared with 51.1--59.8\% for the output features.
Hidden states thus expose skill presence, trajectory matching, and token order under a simple linear readout, whereas the tested output features expose little of any.

\section{Layer-wise Decodability of Matched Hindsight}
\label{app:layerwise_decodability}
\suppressfloats[t]

We expand the matched-versus-other-trajectory comparison in
Figure~\ref{fig:hindsight_decodability} by training a separate
linear probe on each Transformer layer.

The data, trajectory-grouped folds, and output-feature
baselines follow Section~\ref{sec:hindsight_decodability}.
\paragraph{Output-side features probe different views of the
same predictive distribution.}
Let $p_t^{(c)} \in \mathbb{R}^{|\mathcal{V}|}$ denote the
model's next-token probability vector at response position
$t$ under context condition $c$, where $\mathcal{V}$ is
the vocabulary.
The sampled response token $y_t$ is held fixed across
conditions.
We construct three feature representations for separate
logistic regression probes:

\begin{itemize}[leftmargin=*, itemsep=2pt, topsep=2pt]
    \item \textbf{Sampled-token log-probability.}
    The scalar
    \[
        z_{t,\mathrm{sampled}}^{(c)}
        = \log p_t^{(c)}(y_t)
    \]
    records the probability assigned to the token actually
    sampled in the rollout.
    It captures a single coordinate of the output
    distribution, not the probabilities of alternative tokens.

    \item \textbf{Top-64 log-probabilities.}
    Let $K_t$ contain the 64 highest-probability tokens
    under the skill-free context.
    We use
    \[
        z_{t,\mathrm{top64}}^{(c)}
        =
        \bigl(\log p_t^{(c)}(v)\bigr)_{v \in K_t}.
    \]
    The candidate tokens and their ordering are fixed
    across conditions at each response position.
    This feature captures changes among plausible
    alternatives, rather than only the sampled token.

    \item \textbf{256-dimensional distribution summary.}
    We apply a fixed random linear projection
    $P \in \mathbb{R}^{256 \times |\mathcal{V}|}$:
    \[
        z_{t,\mathrm{proj}}^{(c)}
        = P p_t^{(c)}.
    \]
    This feature compresses the full vocabulary
    probability vector into 256 dimensions.
    The same projection is used across conditions;
    each coordinate combines probabilities from the
    vocabulary rather than representing a selected token.
\end{itemize}

For each feature representation, the probe predicts the
context condition, not the next token or task outcome.
Higher classification accuracy indicates that the tested
features make the context distinction more linearly
accessible.
Conversely, near-chance accuracy does not establish that
the full output distribution contains no relevant
information: the selected features, compression, or
linear readout may fail to expose it.

\paragraph{The most decodable layer depends on the backbone.}
Figure~\ref{fig:depth}(c,d) shows that in both models, discrimination is near chance at the earliest
layers and improves with further processing.
For Qwen2.5-3B-Instruct, accuracy reaches 65.4\% at layer 18,
rises further in later intermediate layers, and ends at
66.0\%.
For Qwen3-1.7B-Instruct, accuracy reaches 79.4\% at layer 14 and remains
high across several intermediate layers before declining
to 70.1\% at the final layer.
The middle-layer annotations are reference points, not
claims about the peak accuracy.

\paragraph{Output-feature baselines remain weak for this
specific distinction.}
The three output-feature probes at the right of Figure~\ref{fig:depth}(c,d) achieve approximately
50--52\% accuracy for both backbones, whereas many hidden
layers support substantially stronger discrimination.
This depth profile shows why a final-layer-only diagnostic
can miss variation in the accessibility of hindsight-related
information.

\section{Representational Structure under Hindsight Conditioning}
\label{app:cka_hindsight}

\paragraph{Controlled comparisons characterize the structural
effect of hindsight.}
We examine the initial Qwen2.5-3B-Instruct policy on ALFWorld
using responses captured at rollout step~1, before GRPO
warm-start training.
We hold the model weights and sampled responses fixed and
vary only the preceding context.
The four conditions are a trajectory's self-generated skill,
an other-trajectory skill, a token-shuffled version of the
same skill, and length-matched unrelated filler text.
All conditions are compared with the skill-free context.

Let $H_0^{(\ell)}$ and $H_c^{(\ell)}$ denote the matrices
of hidden states at corresponding response positions in
layer $\ell$, under the skill-free context and condition
$c$, respectively.
We use linear centered kernel alignment (CKA) to compare
their representational similarity structure.
Unlike token-wise cosine distance, CKA compares the
relationships across response positions rather than
the direction of each individual hidden-state vector.
We first examine cross-layer similarity maps and then
resolve same-layer differences across the four conditions.

\begin{figure}[t]
    \centering
    \includegraphics[width=\linewidth]
        {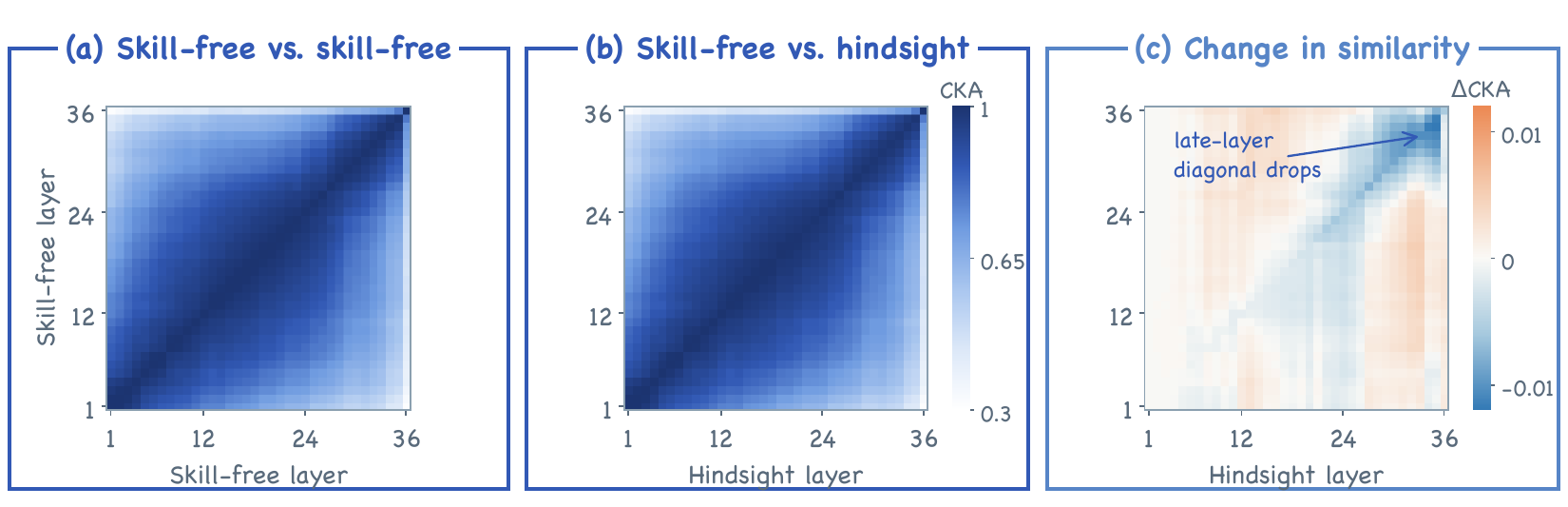}
    \caption{
        \textbf{Cross-layer representational structure under
        hindsight conditioning.}
        Linear CKA on corresponding response positions of
        the initial Qwen2.5-3B-Instruct policy on ALFWorld.
        Model weights and sampled response tokens are fixed.
        (a) Skill-free representations compared across layers:
        $A_{ij}=\mathrm{CKA}(H_0^{(i)},H_0^{(j)})$.
        (b) Skill-free representations compared with those
        conditioned on the trajectory's self-generated skill:
        $B_{ij}=\mathrm{CKA}(H_0^{(i)},H_{\mathrm{own}}^{(j)})$.
        (c) The difference $B-A$, where blue denotes lower
        CKA and orange denotes higher CKA after changing
        the column context.
        Rows index skill-free layers throughout.
        Panels (a,b) share a common color scale; panel (c)
        uses a separate, zero-centered scale.
    }
    \label{fig:cka_layerwise_maps}
\end{figure}

\paragraph{Hindsight largely preserves the global
representational structure.}
Figure~\ref{fig:cka_layerwise_maps} shows closely matching
cross-layer similarity patterns with and without hindsight.
For the self-generated-skill condition, same-layer linear
CKA averages $0.9959$ across the 36 layers and remains
at least $0.9884$.
The difference map nevertheless reveals nonuniform
adjustments, including reduced similarity around the
late-layer diagonal, with a maximum absolute difference
of $0.0118$ across all layer pairs.
Thus, hindsight conditioning changes the representations
while largely preserving their overall cross-layer
similarity structure.
The negative diagonal differences are relative to
self-comparisons of CKA~$1$, not evidence of degraded
task performance.

\paragraph{Same-layer dissimilarity resolves the differences
beneath high CKA.}
To examine these changes more directly, we measure
\begin{equation}
    d_c^{(\ell)}
    =
    1-\operatorname{CKA}
    \bigl(H_0^{(\ell)},H_c^{(\ell)}\bigr).
    \label{eq:cka_dissimilarity}
\end{equation}
Larger values indicate a greater change in representational
similarity structure.
Figure~\ref{fig:depth}(e,f) compares this quantity
across depth using both the linear and debiased CKA
estimates from the diagnostic export.

\paragraph{Self-generated hindsight produces stronger
middle-to-late-layer changes than the controls.}
The self-generated-skill curve rises through much of
the network, reaches its maximum at layer~34, and
declines at the final layer.
From layer~13 onward, its linear-CKA dissimilarity
exceeds that of all three controls.
Averaged across layers, dissimilarity is $0.00406$
for self-generated skills, compared with $0.00269$
for other-trajectory skills, $0.00289$ for shuffled
skills, and $0.00067$ for filler text.
The debiased estimates preserve the same qualitative
separation; the largest absolute difference between
the two estimators is below $0.00067$ across all
conditions and layers.
These comparisons indicate that the structural response
depends on the supplied context and is not reproduced
by the filler control alone.

\paragraph{High global similarity coexists with
context-dependent adjustments.}
Together, the heatmaps and control curves provide
complementary views of hindsight conditioning:
the overall representational structure remains highly
similar, while more detailed comparisons reveal
depth-dependent differences between context conditions.
This complements the token-wise sensitivity and
linear-probe analyses, which examine individual-vector
changes and the accessibility of hindsight distinctions.
The CKA results characterize where structural changes
occur; whether representation supervision improves
policy learning is assessed separately through the
training ablations.

\section{Alignment Dynamics During Training}
\label{app:align_dynamics}

\begin{figure}[!ht]
    \centering
    \includegraphics[width=\linewidth]{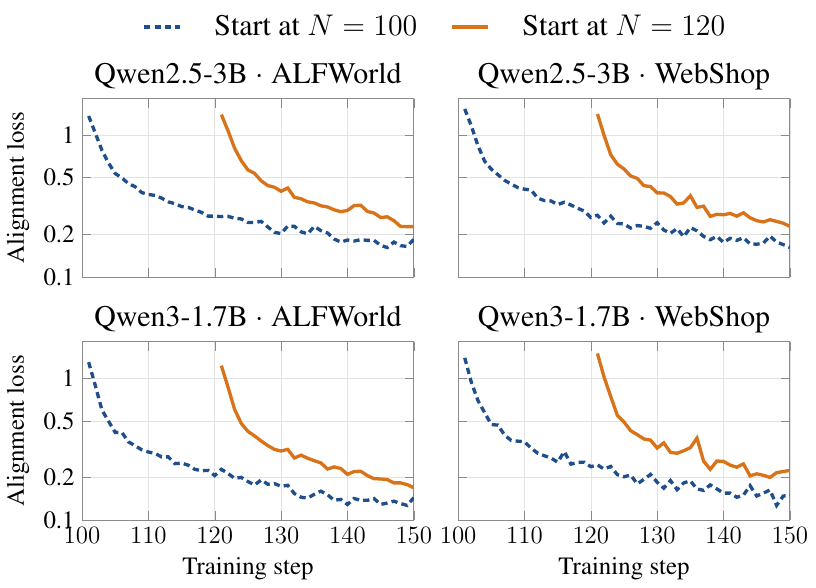}
    \caption{\textbf{Alignment loss of PR-OPD during Stage~II.}
    $\mathcal{L}_{\mathrm{PRA}}$ with the projector at every training step, when representation supervision starts after $N=100$ (dashed) or $N=120$ (solid) GRPO updates, on a logarithmic axis.
    Values are raw per-step training logs of a single run, without smoothing.}
    \label{fig:oprd_loss}
\end{figure}

\textbf{Setup.}
Figure~\ref{fig:oprd_loss} tracks the alignment loss $\mathcal{L}_{\mathrm{PRA}}$ of Eq.~\ref{eq:propd} for PR-OPD with the projector, started after $N=100$ or $N=120$ GRPO updates on both backbones and both environments, and Table~\ref{tab:align_dynamics} summarizes each curve.
The loss is averaged over layers and response tokens and equals $2-2\cos$ between the projected student state and the teacher state~(Appendix~\ref{app:cosine}), so a loss of 0.2 corresponds to a mean cosine similarity of 0.9.
Table~\ref{tab:main} uses $N=100$ on ALFWorld and $N=120$ on WebShop, and every run ends at step 150.

\begin{table}[!ht]
\centering
\caption{Summary of the alignment curves in Figure~\ref{fig:oprd_loss}.
Init.\ is the loss at the first Stage-II update, and ${<}0.3$ is the first update at which the loss falls below 0.3.
The next three columns average five consecutive steps: updates 26--30 compare the two starts after the same number of supervised updates, and steps 146--150 compare them at the end of training.
The last column is the loss of the matching run without the projector, measured on the student's raw hidden states.}
\label{tab:align_dynamics}
\small
\setlength{\tabcolsep}{5pt}
\begin{tabular}{@{}lccccccc@{}}
\toprule
& & & & \multicolumn{3}{c}{\textbf{w/ projector}} & \textbf{w/o proj.} \\
\cmidrule(lr){5-7}\cmidrule(l){8-8}
\textbf{Setting} & \textbf{$N$} & \textbf{Init.} & \textbf{${<}0.3$} & \textbf{Upd.\ 26--30} & \textbf{Steps 146--150} & \textbf{$\cos$} & \textbf{Steps 146--150} \\
\midrule
\multirow{2}{*}{3B $\cdot$ ALFWorld}   & 100 & 1.360 & 17 & 0.224 & 0.170 & 0.915 & 0.012 \\
                                       & 120 & 1.391 & 18 & 0.239 & 0.239 & 0.881 & 0.012 \\
\multirow{2}{*}{3B $\cdot$ WebShop}    & 100 & 1.524 & 19 & 0.228 & 0.175 & 0.913 & 0.008 \\
                                       & 120 & 1.405 & 18 & 0.242 & 0.242 & 0.879 & 0.009 \\
\midrule
\multirow{2}{*}{1.7B $\cdot$ ALFWorld} & 100 & 1.296 & 11 & 0.180 & 0.134 & 0.933 & 0.031 \\
                                       & 120 & 1.224 & 12 & 0.181 & 0.181 & 0.910 & 0.033 \\
\multirow{2}{*}{1.7B $\cdot$ WebShop}  & 100 & 1.389 & 12 & 0.195 & 0.148 & 0.926 & 0.027 \\
                                       & 120 & 1.492 & 13 & 0.213 & 0.213 & 0.893 & 0.026 \\
\bottomrule
\end{tabular}
\end{table}

\paragraph{Alignment is established within a few updates.}
When representation supervision begins, the freshly initialized projector yields a loss of 1.22 to 1.52, a cosine similarity of only 0.24 to 0.39 with the teacher.
Within 4 to 7 updates the loss falls below 0.5, and within 11 to 19 updates below 0.3, so after about ten updates the projected student already points in nearly the same direction as its privileged self.
All eight runs follow this pattern, showing that the alignment objective is easy to optimize across backbones, environments, and starting points.

\paragraph{Alignment improves steadily and stably throughout Stage~II.}
After the fast initial phase, the loss keeps decreasing at a steady pace.
With $N=100$, it drops by a further 23--26\% between updates 26--30 and 46--50 in all four settings and reaches a cosine similarity of 0.913 to 0.933, and it is still decreasing at step 150, leaving room for longer Stage-II training.
No run diverges or degrades late in training.
Even the largest single-step increase after the first ten updates, 0.054 on 1.7B WebShop with $N=120$ at step 136, is fully recovered at the next step.
Because the teacher shares the student's weights, its layers correspond one-to-one with the student's and its target evolves together with the policy, which keeps the objective well behaved throughout training.

\paragraph{Alignment dynamics are robust to the starting point.}
Matched by the number of supervised updates rather than by global step, the two starts follow almost identical trajectories.
Averaged over updates 26--30, their losses differ by at most 0.018 in every setting, and they fall below 0.3 within one update of each other~(Table~\ref{tab:align_dynamics}).
The optimization of PR-OPD is therefore insensitive to when it begins, and the gap between the two curves at step 150 in Figure~\ref{fig:oprd_loss} simply reflects the 20 additional updates of the earlier start.
This lets the warm-start point be chosen for the policy's competence alone, without affecting how well the student can follow its teacher.

\paragraph{A later start turns the same alignment into larger gains.}
On WebShop with 1.7B, starting at $N=120$ raises accuracy from 49.5\% to 57.0\%~(Table~\ref{tab:sensitivity}) with only 30 rather than 50 supervised updates, while its alignment proceeds at the same rate as that of the earlier start.
Each supervised update is thus worth more once the policy has learned the task, in line with the warm-start rationale of Section~\ref{sec:method-overview}: a more competent policy writes more informative hindsight, and PR-OPD transfers it efficiently.
The benefit of PR-OPD therefore lies in what the privileged teacher knows rather than in driving the alignment loss to zero, and 30 updates of representation supervision suffice for the WebShop gains in Table~\ref{tab:main}.

\paragraph{The behavior holds across backbones and environments.}
The same fast-then-steady pattern appears on both backbones and both environments.
The smaller backbone aligns even more readily: 1.7B falls below 0.3 after 11 to 13 updates, against 17 to 19 for 3B, and ends with a lower loss under every environment and start, for example 0.134 and 0.148 against 0.170 and 0.175 with $N=100$.

\paragraph{The projector turns a fixed gap into a learnable target.}
Without the projector, the loss is measured directly on the student's raw hidden states and is not comparable to the projected loss.
It is 0.008 to 0.012 for 3B and 0.026 to 0.033 for 1.7B over the last five steps and stays within 0.01 of its initial value in every run, because the teacher shares the student's weights and moves with every update.
The projector instead gives the student a mapping it can learn, so the alignment loss decreases steadily while the raw states that feed the language-model head remain free to serve the policy~(Section~\ref{sec:propd}).
The raw gap is also 2.5 to 3.5 times larger for 1.7B than for 3B, in line with the higher probe accuracy of 1.7B in Section~\ref{sec:general_analysis}, and 1.7B is exactly where the projector matters most~(Section~\ref{sec:ablation}), lifting ALFWorld success from 36.7\% to 46.9\%.

Together, these curves show that the representation objective of PR-OPD is fast to fit, stable over training, consistent across settings, and insensitive to when it starts, so its gains come from the privileged hindsight it transfers rather than from delicate optimization.